\documentclass[a4paper,11pt]{article}
\usepackage{amsmath}
\usepackage{amssymb}
\usepackage[pdftex]{graphicx}
\usepackage{epstopdf}
\usepackage{longtable}
\usepackage{subfigure}
\usepackage{natbib}
\usepackage[pdftex, pdftitle={The Socceral Force},
pdfauthor={Norbert Batfai},
pdfkeywords={simulation of football, XML-based modeling, Football World Cup, mobile game, RELAX NG compact syntax, Java, open source},
pdfstartview=FitB]{hyperref}
\usepackage{amssymb,amsmath,amsthm,amsfonts}

\newcommand{\keywords}[1]{\par\addvspace\baselineskip\noindent\textbf{Keywords: }\textit{#1}.}

\author{Norbert B\'atfai
\\University of Debrecen
\\Department of Information Technology
\\\texttt{batfai.norbert@inf.unideb.hu}}
\title{The Socceral Force}
\begin{document}
\maketitle

\begin{abstract}
We have an audacious dream, we would like to develop a simulation and virtual
reality system to support the decision making in European football (soccer). 
In this review, we summarize the efforts that we have made to fulfil this dream until recently.
In addition, an introductory version of FerSML (Footballer and Football Simulation Markup Language)
is presented in this paper.
\end{abstract}

\tableofcontents

\section{Introduction}

We are working to develop an expert system that can support decision making in European football (soccer). Based on the observation of players and coaches, it will be developed as an overlay system to our soccer simulation model. In theory, we are supposed to be able to take a comprehensive look at how the players behave in the field. There are already several (GPS and video based  \citep{Brillinger}, \citep{gps}, \citep{19897414}, \citep{BarberoAlvarez2010232}) methods for this. But in case of coaches, to introduce computerization and automation seems to be a more difficult problem, however the behavior of the coaches may be simply analyzed by questionnaires. By our terminology, the avatar is a set of observed data which enables us to give some probable answer to behavior issues of footballers and coaches. The observed properties of players and coaches and the control of simulations have also been described in a new open source XML language called Footballer and Football Simulation Markup Language, or briefly FerSML.

In choosing the title for this paper we were inspired by \citep{citeulike:6677720}.  In this cited study a short part of a match of the World Cup 2006 (Argentina 25 passes goal)  was analyzed  by building a potential field based on a regression model.

The next subsection of introduction contains our previous work. The second section gives a complete review of the introductory version of FerSML. The third section shows the software which are being developed by us. In its subsection, some early simulations are also shown and we will point to some criteria to be used when selecting the appropriate simulation model.

\subsection{Previous Work}

Our idea was introduced in \citep{batfai-jcscs}. Here we give an introduction to planning a new XML language called FerSML (\textit{Football(er) Simulation Markup Language}).  It is a language for describing the footballers, the head coaches and the simulations themselves. 

\subsubsection{The \textit{Soccer Game 4u OSE} Mobile Game}

\begin{figure}[h!]
\centering
\subfigure[The USA team celebrate a goal.]{\includegraphics[scale=0.6]{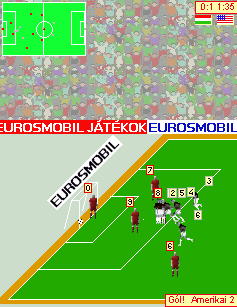}}
\subfigure[An American corner kick.]{\includegraphics[scale=0.6]{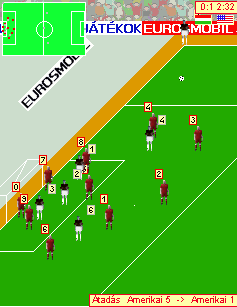}}
\subfigure[The playing systems 4-3-2 and 3-4-2.]{\includegraphics[scale=0.6]{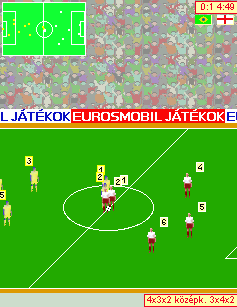}}\\
\subfigure[The German team is building an attack.]{\includegraphics[scale=0.6]{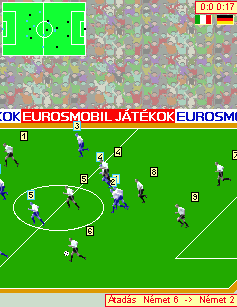}}
\subfigure[The German team won the match with 0:2.]{\includegraphics[scale=0.6]{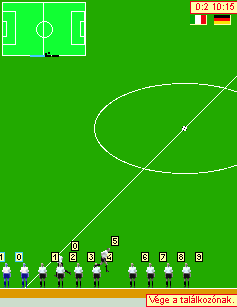}}
\subfigure[A kick from the Penalty Mark.]{\includegraphics[scale=0.6]{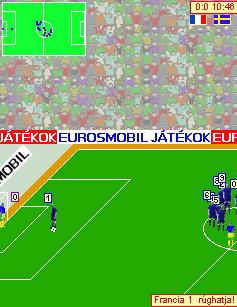}}
\caption{Screenshots from the \textit{Soccer Game 4u OSE} game.\label{sg4u}}
\end{figure}

At this moment the FerSML simulation is entirely based on the soccer simulator built in Eurosmobil's mobile soccer games. These were opened by Eurosmobil in the framework of the author's PhD dissertation \citep{norbi-phd}. The mobile games in question are released in SourceForge.net (\url{https://sourceforge.net/projects/javacska/}) under the project name \textit{Javacska One (J\'av\'acska One)}. Here the name of the soccer game is \textit{Soccer Game 4u OSE}, or in Hungarian \textit{Focij\'at\'ek Neked NYFK}. It is protected by the GNU General Public License version 3 (GNU GPL v3). Some screenshots of this game are shown in Figure\ref{sg4u}. The open source games of \textit{J\'av\'acska One} have been used in higher and secondary education \citep{icai}, \citep{tmcs}.

\subsubsection{The FerSML project}

Turning to the case in FerSML, this project is also maintained in SourceForge.net (\url{https://sourceforge.net/projects/footballerml/}).  It is natural that the process of the filling of avatar files is closely connected with football clubs, because the data in question shall be confidential. But it is interesting to note that this does not contest that the structure of avatar files and the simulation program are released under an open source license. This will enable hardcore football fans to participate in the project. For example, they are able to create their own avatars of their favourite teams and to run their own simulations with our FerSML software.

\subsection{Related work}

There are many articles in literature which study the relationship between football and modelling of football  \citep{Brillinger}, \citep{DBLP:journals/eor/KoningKRR03}, \citep{citeulike:6677405}, \citep{citeulike:6677720}. But at the time of writing, we have no information as to whether the simulation of the world of playing football was described in XML.

\section{Footballer and Football Simulation Markup Language}

\begin{figure}[h!]
\begin{center}
\includegraphics[scale=0.6]{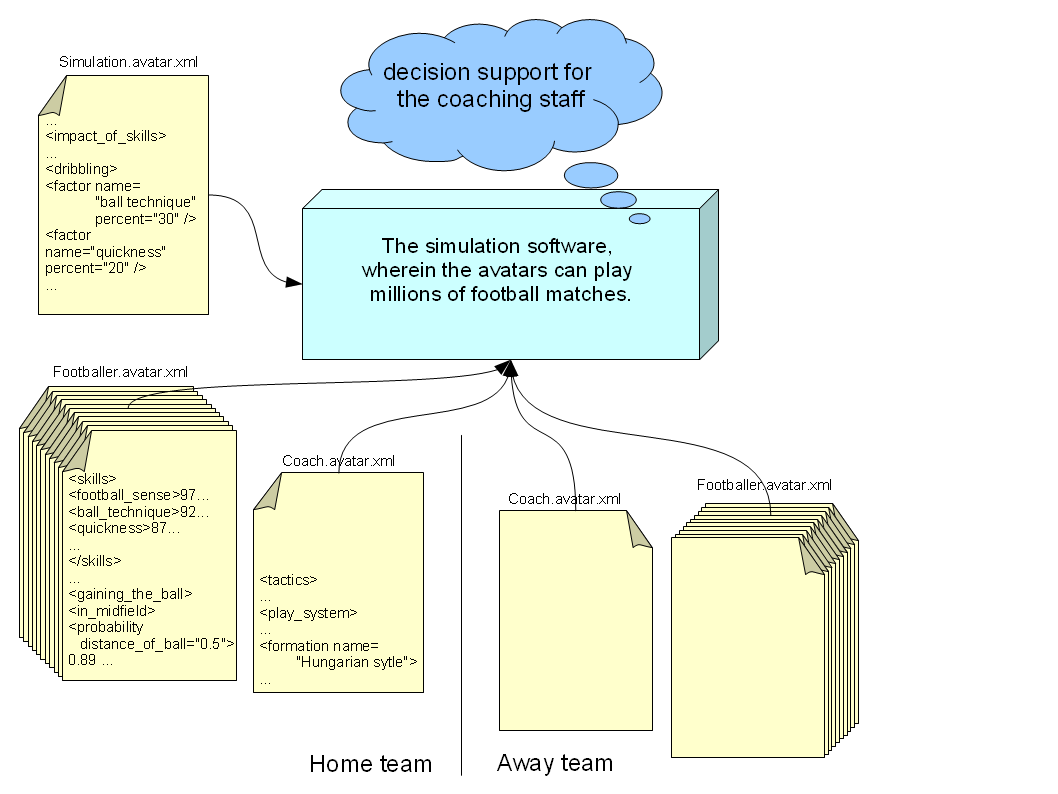}
\caption{The structure of using FerSML platform.\label{fersml}}
\end{center}
\end{figure}

The Figure\ref{fersml} depictes the structure of using FerSML platform. 

\subsection{Football(er) Simulation Markup Language}

The open structure of avatars is described in Relax NG compact syntax (RNC, \url{http://relaxng.org/}). 
A sample XML and the RNC file are presented in the following subsections.

\subsubsection{A FerSML.avatar.xml File}

\begin{verbatim}
<?xml version="1.0" encoding="UTF-8"?>
<!--
 # FerSML.avatar.xml
 #
 # Football(er) Simulation Markup Language
 #
 # Copyright (C) 2010, Batfai Norbert
 #
 # This program is free software: you can redistribute it and/or modify
 # it under the terms of the GNU General Public License as published by
 # the Free Software Foundation, either version 3 of the License, or
 # (at your option) any later version.
 #
 # This program is distributed in the hope that it will be useful,
 # but WITHOUT ANY WARRANTY; without even the implied warranty of
 # MERCHANTABILITY or FITNESS FOR A PARTICULAR PURPOSE.  See the
 # GNU General Public License for more details.
 #
 # You should have received a copy of the GNU General Public License
 # along with this program.  If not, see <http://www.gnu.org/licenses/>.
 # -
 # FerSML, Football(er) Simulation Markup Language
 # https://sourceforge.net/projects/footballerml/
 #
 # Further information can be found in http://arxiv.org/abs/1004.2003
 # -
 # FerSML
 # 2010.04.12.
 # -
 #
 # Version history:
 #
 # 0.0.2     an introductory version for testing purposes
 #
 -->
<fersml>
    <coach>
        <starting_team>
            <player player_id="1" squad_number="9" />
            <player player_id="2" squad_number="19" />
            <player player_id="3" squad_number="10" formation_name="4-4-2" />
            <player player_id="3" squad_number="9" formation_name="4-3-3" />
            <player player_id="4" squad_number="39" />
            <player player_id="5" squad_number="49" />
            <player player_id="6" squad_number="59" />
            <player player_id="7" squad_number="69" />
            <player player_id="8" squad_number="79" />
            <player player_id="9" squad_number="89" />
            <player player_id="10" squad_number="99" />
            <player player_id="11" squad_number="8" />
        </starting_team>
    </coach>
    <avatar>
        <person squad_number="99">
            <firstname>Firstname</firstname>
            <lastname>Lastname</lastname>
            <age>99</age>
            <height>99</height>
            <weight>99</weight>
            <dominant_foot>both</dominant_foot>
            <usual_position>
              attacking midfielder
            </usual_position>
            <actual_position>
              left winger
            </actual_position>
        </person>
        <estimations>
            <skills>
                <football_sense>97</football_sense>
                <ball_technique>92</ball_technique>
                <quickness>87</quickness>
            </skills>
            <actions>
                <shutting_goal>
                    <prob dist="5">
                      0.89
                    </prob>
                    <prob dist="16">
                      0.84
                    </prob>
                    <prob dist="30">
                      0.47
                    </prob>
                </shutting_goal>
                <gaining_ball>
                    <prob dist="0.5">
                      0.89
                    </prob>
                    <prob dist="1">
                      0.64
                    </prob>
                    <prob dist="2">
                      0.06
                    </prob>
                </gaining_ball>
            </actions>
        </estimations>
    </avatar>
    <simulation>
        <control>
            <impact_of_skills>
                <dribbling>
                    <factor name="ball technique"
                        percent="30" />
                    <factor name="quickness"
                        percent="20" />
                </dribbling>
                <shielding>
                    <factor name="football sense"
                        percent="30" />
                    <factor name="ball technique"
                        percent="30" />
                    <factor name="quickness"
                        percent="20" />
                </shielding>
                <tackling>
                    <factor name="ball technique"
                        percent="20" />
                    <factor name="quickness"
                        percent="30" />
                </tackling>
            </impact_of_skills>
        </control>
        <knowledge_base>
            <tactics>
                <play_system>
                    <!-- The 3-3-3 play system -->
                    <formation name="3-3-3">
                        <!-- Goalkeeper -->
                        <player_position player_id="1" desc="keeper">
                            <coord_x>10</coord_x>
                            <coord_y>320</coord_y>
                        </player_position>
                        <!-- Defenders -->
                        <player_position player_id="9" desc="defender">
                            <coord_x>845</coord_x>
                            <coord_y>470</coord_y>
                        </player_position>
                        <player_position player_id="8" desc="central defender">
                            <coord_x>860</coord_x>
                            <coord_y>320</coord_y>
                        </player_position>
                        <player_position player_id="7" desc="defender">
                            <coord_x>835</coord_x>
                            <coord_y>230</coord_y>
                        </player_position>
                        <!-- Midfielders -->
                        <player_position player_id="6">
                            <coord_x>640</coord_x>
                            <coord_y>530</coord_y>
                        </player_position>
                        <player_position player_id="5">
                            <coord_x>650</coord_x>
                            <coord_y>310</coord_y>
                        </player_position>
                        <player_position player_id="4">
                            <coord_x>640</coord_x>
                            <coord_y>100</coord_y>
                        </player_position>
                        <!-- Attackers -->
                        <player_position player_id="3">
                            <coord_x>410</coord_x>
                            <coord_y>400</coord_y>
                        </player_position>
                        <player_position player_id="2">
                            <coord_x>470</coord_x>
                            <coord_y>330</coord_y>
                        </player_position>
                        <player_position player_id="10">
                            <coord_x>410</coord_x>
                            <coord_y>240</coord_y>
                        </player_position>
                    </formation>
                </play_system>
            </tactics>
        </knowledge_base>    
    </simulation>
</fersml>
\end{verbatim}

\subsubsection{The FerSML.avatar.rnc File}

\begin{verbatim}
 # FerSML.avatar.rnc
 #
 # Football(er) Simulation Markup Language
 #
 # Copyright (C) 2010, Batfai Norbert
 #
 # This program is free software: you can redistribute it and/or modify
 # it under the terms of the GNU General Public License as published by
 # the Free Software Foundation, either version 3 of the License, or
 # (at your option) any later version.
 #
 # This program is distributed in the hope that it will be useful,
 # but WITHOUT ANY WARRANTY; without even the implied warranty of
 # MERCHANTABILITY or FITNESS FOR A PARTICULAR PURPOSE.  See the
 # GNU General Public License for more details.
 #
 # You should have received a copy of the GNU General Public License
 # along with this program.  If not, see <http://www.gnu.org/licenses/>.
 # -
 # FerSML, Football(er) Simulation Markup Language
 # https://sourceforge.net/projects/footballerml/
 #
 # Further information can be found in http://arxiv.org/abs/1004.2003 
 # -
 # FerSML
 # 2010.04.12.
 # -
 #
 # Version history:
 #
 # 0.0.2     an introductory version for testing purposes
 #
grammar {
    start = FerSML
    FerSML = 
    element fersml {
        Coach, 
        Avatar*, 
        Simulation
    }
    Coach = 
    element coach {
       element starting_team {
            element player {
                attribute player_id {
                    xsd:integer {
                        minInclusive = "1"
                        maxInclusive = "11"
                    }
                },
                attribute squad_number {
                    xsd:integer {
                        minInclusive = "0"
                        maxInclusive = "99"
                    }
                },
                attribute formation_name {
                    text
                }?        
            }*
        }
    }        
    Avatar = 
    element avatar {
        element person {
            attribute squad_number {
                xsd:integer {
                    minInclusive = "0"
                    maxInclusive = "99"
                }
            },
            element firstname {
                text
            },    
            element lastname {
                text
            },    
            element age {
                xsd:positiveInteger
            },        
            element height {
                xsd:positiveInteger
            },    
            element weight {
                xsd:positiveInteger
            },        
            element dominant_foot {
                "both"
                | "left"
                | "right"
            },    
            element usual_position {
                Positions
            },    
            element actual_position {
                Positions
            }           
        },
        element estimations {
            element skills {
                element football_sense {
                    xsd:integer {
                        minInclusive = "1"
                        maxInclusive = "100"
                    }
                },            
                element ball_technique {
                    xsd:integer {
                        minInclusive = "1"
                        maxInclusive = "100"
                    }
                },            
                element quickness {
                    xsd:integer {
                        minInclusive = "1"
                        maxInclusive = "100"
                    }
                }                    
            },
            element actions {
                element shutting_goal {
                    element prob {
                        attribute dist {
                            xsd:float {
                                minInclusive = "0.00"
                                maxInclusive = "1024.00"
                            }
                        },
                        xsd:float {
                            minInclusive = "0.00"
                            maxInclusive = "1.00"                    
                        }
                    }*                        
                }?,        
                element gaining_ball {
                    element prob {
                        attribute dist {
                            xsd:float {
                                minInclusive = "0.00"
                                maxInclusive = "1024.00"
                            }
                        },
                        xsd:float {
                            minInclusive = "0.00"
                            maxInclusive = "1.00"                    
                        }
                    }*                        
                }?        
            }        
        }    
    }
    Simulation =
    element simulation {
        element control {
            element impact_of_skills {
                element dribbling {      
                    element factor {
                        attribute name { text },
                        attribute percent { xsd:integer {
                            minInclusive = "1" 
                            maxInclusive = "100" }				
                        }
                    }*
                },
                element shielding {
                    element factor {
                        attribute name { text },
                        attribute percent { xsd:integer {
                            minInclusive = "1" 
                            maxInclusive = "100" }								
                        }
                    }*
                },
                element tackling {		
                    element factor {
                        attribute name { text },
                        attribute percent { xsd:integer {
                            minInclusive = "1" 
                            maxInclusive = "100" }								
                        }
                    }*
                }
            }
        },
        element knowledge_base {
            element tactics {
                element play_system {
                    element formation {
                        attribute name {
                            text
                        },
                        element player_position {
                            attribute player_id {
                                xsd:integer {
                                    minInclusive = "1"
                                    maxInclusive = "11"
                                }
                            },
                            attribute desc {
                                Positions
                            }?,
                            element coord_x {
                                xsd:integer {
                                    minInclusive = "0"
                                    maxInclusive = "1024"
                                }
                            },            
                            element coord_y {
                                xsd:integer {
                                    minInclusive = "0"
                                    maxInclusive = "640"
                                }
                            }
                        }*
                    }+
                }
            }
        }
    }    
    Positions = 
        "keeper"
        | "midfielder"
        | "defensive midfielder"
        | "attacking midfielder"
        | "winger"
        | "left winger"
        | "right winger"
        | "forward"    
        | "deep-lying forward"
        | "centre forward"
        | "striker"
        | "inside forward"                            
        | "playmaker"
        | "sweeper"
        | "defender"
        | "central defender"
        | "centre back"
        | "wing back"
        | "full fback"
        | "half back" 
}
\end{verbatim}

\subsection{The Development of the Related  Simulation Software}

A legacy of the \textit{Soccer Game 4u OSE} is that a team consists of only 10 players because the players were represented by the hard key buttons of the phone.

\subsubsection{Automated Soccer Applet for FerSML}

This program helps to visualize the tested simulation models. In our terminology, it helps to study the \textit{criterion TV}. It is met if the flow of play seems football rather than tennis, ping pong or rugby. This criterion is used first when we are searching for such simulation models in which the distribution of real and simulated phenomena (for example, the total number of goals) are equal.

\begin{figure}[h!]
\begin{center}
\includegraphics[scale=0.6]{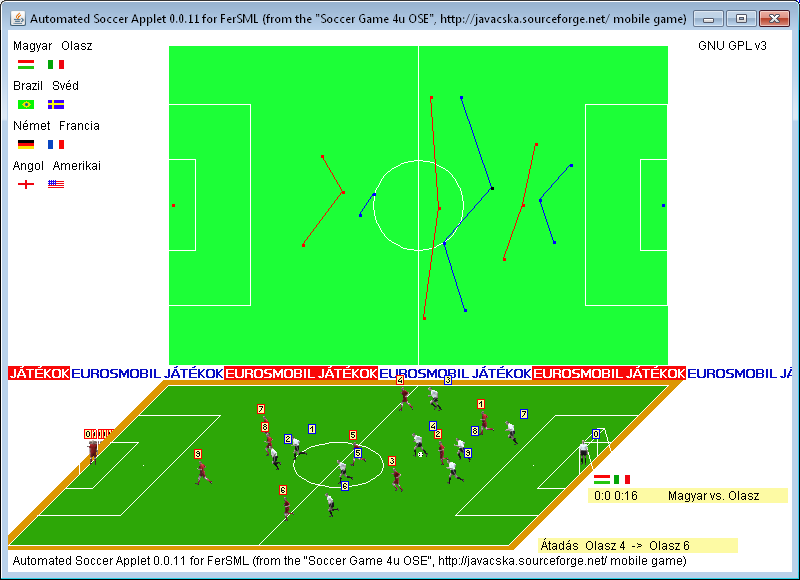}
\caption{A screenshot from the \textit{Automated Soccer Applet for FerSML}.\label{asa}}
\end{center}
\end{figure}

\paragraph{Real and simulated  Football World Cups}

\begin{table}[h!]
\begin{center}
\begin{tabular}{|c|c|c|c|c|c|c|c|c|c|}
  \hline
  1934&1954&1958&1966&1970&1990&1994&1998&2002&2006 \\
  \hline
  27&48&35&31&34&30&33&27&22&29\\
  \hline
\end{tabular}
\caption{The number of goals in the quarter-finals, semi-finals, match for third place and final of the 10 selected Football World Cups.\label{table10}}
\end{center}
\end{table}

Table \ref{table10} shows the number of total goals in the quarter-finals, semi-finals, match for third place and final of the 10 selected Football World Cups \citep{fifawc}.

\begin{table}[h!]
\begin{center}
\begin{tabular}{|c|c|c|c|c|c|c|c|c|c|c|c|c|}\hline
&1 sc WC&2.&3.&4.&5.&6.&7.&8.&9.&10.&$\overline{x}$&$s_n^*$\\ \hline
1.&35&27&31&34&30&22&29&31&31&31&31,6&3,63\\ \hline
2.&30&26&28&39&29&44&34&29&37&40&33,6&6,09\\ \hline
3.&33&30&24&27&15&26&30&29&28&31&27,3&5,03\\ \hline
4.&26&33&38&23&21&27&31&24&36&35&29,4&5,98\\ \hline
5.&35&27&34&33&31&24&40&32&27&32&31,5&4,60\\ \hline
6.&30&37&26&29&33&33&29&23&32&24&29,6&4,37\\ \hline
7.&31&23&22&32&31&27&26&18&29&38&27,7&5,77\\ \hline
8.&26&36&35&22&32&14&28&33&28&30&28,4&6,06\\ \hline
\end{tabular}
\caption{The number of goals in the quarter-finals, semi-finals, match for third place and final of the simulated Football World Cups with 8 different settings of parameters. (The parameter settings are the same in the rows.)\label{table80}}
\end{center}
\end{table}

Table \ref{table80} shows the number of total goals in the quarter-finals, semi-finals, match for third place and final of the  $8\times10$  simulated Football World Cups with $8$ different settings of parameters.

The null hypothesis that the two distributions are identical must not be rejected by the well known Wald-Wolfowitz and Mann-Whitney tests \citep{norbi-ht}.

\subsubsection{Socceral Force Applet for FerSML}

\begin{figure}[h!]
\begin{center}
\includegraphics[scale=0.6]{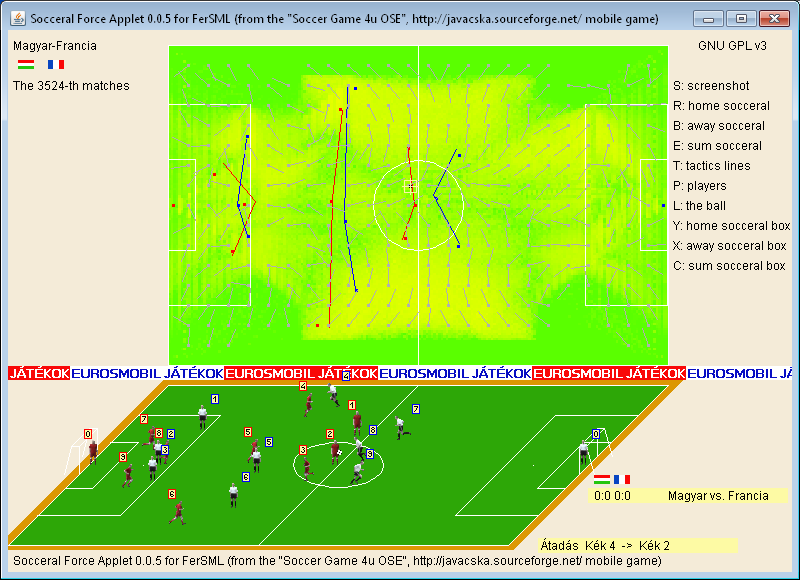}
\caption{A screenshot from the \textit{Socceral Force Applet for FerSML}.\label{sfa}}
\end{center}
\end{figure}

\begin{figure}[h!]
\centering
\subfigure[The socceral force of the away team.]{\includegraphics[scale=0.7]{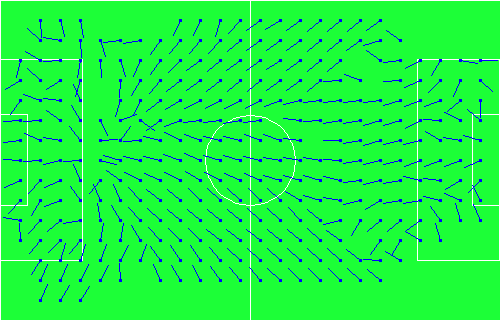}}
\subfigure[The socceral force of the home team.]{\includegraphics[scale=0.7]{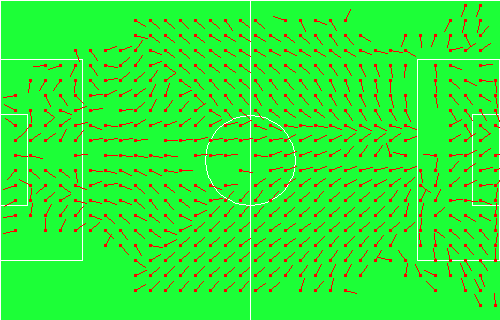}}
\caption{Screenshots from the \textit{Soccer Game 4u OSE} game.\label{sfi}}
\end{figure}

\begin{figure}[h!]
\begin{center}
\includegraphics[scale=0.7]{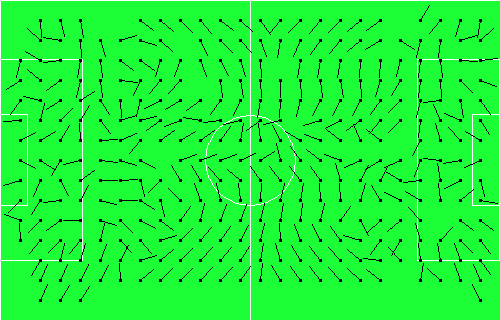}
\caption{The sum of two (home and away socceral force) vector spaces in the \textit{Socceral Force Applet for FerSML}.\label{sfae}}
\end{center}
\end{figure}

Figure \ref{sfa} shows a screenshot of \textit{Socceral Force Applet}. This is a further development of the \textit{Automated Soccer Applet}. It also works on a discrete time scale and stores the movements of the ball in all moments. To be more precise, the ball has an actual position $(lx, ly)$ and a target position  $(lcx, lcy)$ and the next code snippet will run every $100$ milliseconds

\begin{verbatim}
  socceralField[lx][ly][0][kiLep] = 
    (socceralField[lx][ly][0][kiLep] + lcx - lx) / 2;
  socceralField[lx][ly][1][kiLep] = 
    (socceralField[lx][ly][1][kiLep] + lcy - ly) / 2;
\end{verbatim}

where the variable \texttt{kiLep} denotes the home or away team. This \texttt{soccer\-al\-Field} arrays are shown in Figure \ref{sfi}. In addition, the sum of these two sets of vectors is shown in Figure \ref{sfae}. Figure \ref{sfi} also indicates the magnitude of vectors of this sum. The precise details can be found in file \texttt{FootballMatch.java}. 

This further development of \textit{Automated Soccer Applet} was inspired by figure 6, 7 and 8 of \citep{citeulike:6677720}. 

\begin{figure}[h!]
\begin{center}
\includegraphics[scale=0.7]{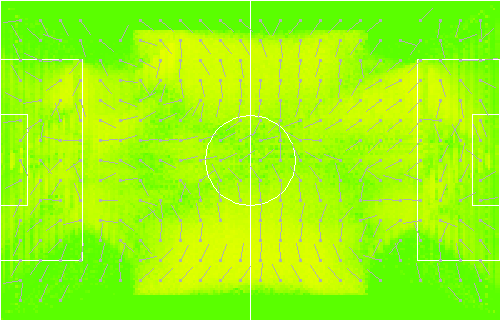}
\caption{The magnitude of vectors is indicated by \texttt{Color(255 * n / N,  255, 0)} where $n$ denotes the length of a vector.\label{sfb}}
\end{center}
\end{figure}

\section{Conclusion and further work}

At the moment we are working to improve our simulations of playing soccer. In parallel with this, we are searching 
appropriate statistical tests and data to ensure that our simulated world and the world of real football are not very different.

\paragraph{Acknowledgements.}
The author would like to thank Professor Gy\"orgy Terdik for drawing his attention to the literature \citep{Brillinger} and \citep{citeulike:6677720} on analysis of football.

\bibliography{TheSocceralForce}

\end{document}